\documentclass[journal]{IEEEtran}
\ifCLASSINFOpdf
\else
\fi
\usepackage{amsmath}
\usepackage{graphicx}
\usepackage{mdwtab}
\usepackage{booktabs}
\usepackage{multirow}
\usepackage{multicol}
\usepackage{xcolor}
\usepackage{amssymb}
\usepackage{hyperref}
\usepackage[ruled,vlined]{algorithm2e}
\usepackage[caption=false,font=normalsize,labelfont=sf,textfont=sf]{subfig}
\usepackage{textcomp}
\usepackage{stfloats}
\usepackage{adjustbox}

\newcommand\Y{\mathbf Y}
\newcommand\E{\mathbf E}
\newcommand\A{\mathbf A}

\newcommand\D{\mathbf D}

\newcommand\B{\mathbf B}

\newcommand\bS{\mathbf S}
\newcommand\bL{\mathbf L}

\newcommand\I{\mathbf I}
\newcommand\Q{\mathbf Q}

\newcommand\1{\bf 1}

\newcommand\yy{\mathbf y}
\newcommand\eg{{\it{e.g. }}}
\newcommand\ie{{\it{i.e. }}}

\newcommand{\Lagr}{\mathcal{L}}
\newcommand\blue{\textcolor{black}}
\begin{document}
%
\title{Fast Semi-supervised Unmixing using Non-convex Optimization} 
%
%
%

\author{Behnood~Rasti,~\IEEEmembership{Senior~Member,~IEEE,}
Alexandre~Zouaoui,~\IEEEmembership{Member,~IEEE,} Julien~Mairal,~\IEEEmembership{Senior Member,~IEEE} and  Jocelyn Chanussot, ~\IEEEmembership{Fellow,~IEEE}
\thanks{Behnood Rasti (corresponding author) is with the Faculty of Electrical Engineering and Computer Science, Technische Universität Berlin, and the Berlin Institute for the Foundations of Learning and Data, Berlin, Germany, behnood.rasti@tu-berlin.de, behnood.rasti@gmail.com. }
\thanks{Alexandre Zouaoui, Jocelyn Chanussot, and Julien Mairal are with Univ. Grenoble Alpes, Inria, CNRS, Grenoble INP, LJK, 38000 Grenoble, France.}
\thanks{Alexandre Zouaoui is currently with Data Science Experts, Grenoble, France.}
\thanks{Manuscript received  2024; revised 2024.}}

%
%

\markboth{Journal of \LaTeX\ Class Files,~Vol.~?, No.~?, ?~2024}%
{Shell \MakeLowercase{\textit{et al.}}: Bare Demo of IEEEtran.cls for Journals}
%



\maketitle

\begin{abstract}

In this paper, we introduce a novel linear model tailored for semisupervised/library-based unmixing. Our model incorporates considerations for library mismatch while enabling the enforcement of the abundance sum-to-one constraint (ASC). Unlike conventional sparse unmixing methods, this model involves nonconvex optimization, presenting significant computational challenges. We demonstrate the efficacy of Alternating Methods of Multipliers (ADMM) in cyclically solving these intricate problems. We propose two semisupervised unmixing approaches, each relying on distinct priors applied to the new model in addition to the ASC: sparsity prior and convexity constraint. Our experimental results validate that enforcing the convexity constraint outperforms the sparsity prior for the endmember library. These results are corroborated across three simulated datasets (accounting for spectral variability and varying pixel purity levels) and the Cuprite dataset. Additionally, our comparison with conventional sparse unmixing methods showcases considerable advantages of our proposed model, which entails nonconvex optimization. Notably, our implementations of the proposed algorithms—fast semisupervised unmixing (FaSUn) and sparse unmixing using soft-shrinkage (SUnS)—prove considerably more efficient than traditional sparse unmixing methods. SUnS and FaSUn were implemented using PyTorch and provided in a dedicated Python package called Fast Semisupervised Unmixing (FUnmix), which is open-source and available at \href{https://github.com/BehnoodRasti/FUnmix}{https://github.com/BehnoodRasti/FUnmix}.

\end{abstract}


%
\IEEEpeerreviewmaketitle

\section{Introduction}

\IEEEPARstart{S}{pectral} unmixing estimates the abundances of pure spectra of materials called endmembers. Depending on the prior knowledge of endmembers, the unmixing problem can be categorized into three main types: supervised unmixing, blind unmixing, and semi-supervised unmixing. In supervised unmixing, abundances are estimated relying on known endmembers. Blind unmixing estimates both endmembers and abundances simultaneously. Semi-supervised unmixing relies on an endmember library to estimate the corresponding abundances \cite{HySUPP}. 

An unmixing problem can be tackled using a sequential process of extracting/estimating endmembers (often using a geometrical approach) followed by an abundance estimation technique. We call this group of methods supervised since the endmembers are assumed to be known in the abundance estimating step \cite{HySUPP}.  Endmember extraction/estimation techniques often rely on the geometry of the data simplex, such as Vertex Component Analysis (VCA) \cite{VCA}, Simplex volume maximization (SiVM) \cite{RHeylen_2011}, the minimum volume simplex analysis (MVSA) \cite{MVSA}, and the simplex identification via variable splitting and augmented Lagrangian (SISAL) \cite{SISAL}. The discussion on the endmember extraction/estimation techniques is out of the scope of this paper, and therefore, we refer to \cite{unmixing-review, HySUPP} for an overview of this topic. 

When the endmembers are known, abundances can be estimated. Unconstrained least squares unmixing (UCLSU) via the orthogonal subspace projection was proposed for abundance estimation \cite{ULS_CI}. Non-negative constrained least squares unmixing (NCLSU) \cite{NCLS, NCLSU} was proposed to estimate the abundances subjected to ANC. There are several attempts to solve the least squares problem subjected to both ANC and ASC \cite{CLS, WLS}. The first efficient algorithm was proposed in \cite{FCLSU} and called fully constrained least squares unmixing (FCLSU). FCLSU can be efficiently solved using general-purpose convex optimization toolboxes. However, in this paper,  we will show that with the advances in graphical processing units (GPU), FCLSU can be efficiently solved using convex optimization techniques i.e., Alternating Direction Method of Multipliers (ADMM) \cite{ADMM}.

The pioneer semisupervised unmixing is Multiple Endmember Spectral and Mixture Analysis (MESMA) \cite{MESMA_Lib_EE} proposed to address endmember variability. MESMA assumes a structured library containing endmember bundles for all materials, allowing different scaled endmembers for each pixel. However, this is a combinatorially complex task and computationally expensive. J. M. Bioucas Dias and M. A. T. Figueiredo \cite{SUnSAL} proposed a formulation of the semi-supervised unmixing problem as a sparse regression problem, giving rise to what is known as sparse unmixing. They proposed sparse unmixing by variable splitting and augmented lagrangian (SUnSAL), as well as its variant known as Constrained SUnSAL (C-SUnSAL) \cite{SUnSAL}. SUnSAL and C-SUnSAL employ the $\ell_1$ penalty to encourage sparsity in the abundance estimation. In the case of SUnSAL, the $\ell_2$ norm is combined with the $\ell_1$ penalty to enhance fidelity, whereas C-SUnSAL uses the $\ell_2$ norm as a constraint in order to minimize the $\ell_1$ term. The optimization problems associated with SUnSAL were tackled through the Alternating Direction Method of Multipliers (ADMM) \cite{ADMM}.

Sparse unmixing offers computational efficiency; however, the high correlation among library endmembers presents a significant challenge for sparse regression. This concern has been addressed through techniques like library pruning and the application of various sparsifying regularizers. To enhance SUnSAL, a total variation (TV) penalty was incorporated, resulting in SUnSAL-TV \cite{SUnSAL-TV}, which leverages spatial information. Nevertheless, SUnSAL-TV does not satisfy the constraint of abundance sum-to-one (ASC) due to its potential conflict with the $\ell_1$ penalty.

Collaborative sparse unmixing \cite{CoSUn} enforces a constraint by applying the sum of $\ell_2$ norms to the abundances. Double Reweighted Sparse Unmixing (DRSU) \cite{DRSU} and 
Spectral-Spatial Weighted Sparse Unmixing (S2WSU) \cite{SZhang2018} adopt a weighted $\ell_1$ norm approach to induce sparsity. Additionally, DRSU employs total variation to capture spatial attributes. 
The Multiscale Sparse Unmixing Algorithm (MUA) \cite{RABorsoi2019} capitalizes on spatial correlation by performing sparse regression on segmented pixels, enabling the capture of both spectral variability and spatial correlation. In this context, segmentation techniques such as Binary Partition Tree (BPT), Simple Linear Iterative Clustering (SLIC), and the K-means algorithm were suggested in \cite{RABorsoi2019}. In \cite{TanerInce2020}, SLIC was adopted for segmentation, while sparse unmixing was executed with superpixel-based graph Laplacian regularization.

A common drawback of the aforementioned sparse unmixing techniques is that the estimated fractional abundances may not accurately represent the aerial fraction of each pure material on the ground due to the absence of the ASC constraint. As mentioned, applying $\ell_1$ penalties to the abundances cannot maintain the ASC. This issue was addressed in sparse unmixing using a convolutional neural network (SUnCNN) \cite{SUnCNN}. In \cite{SUnCNN}, we demonstrated that selecting an appropriate prior for sparse regression could be transformed into an optimization task involving the parameters of a deep encoder-decoder network, while the ASC could be enforced using a softmax layer. However, it is worth noting that selecting suitable hyperparameters for such a deep network is often a challenging endeavor. In \cite{SMALU}, an asymmetric encoder-decoder network is used with a sparse variation of softmax to avoid the full support of softmax while enforcing ASC. 

Algorithm unrolling-based strategies have also been explored in the context of sparse unmixing. A recent instance of this approach involves the development of a shallow network for sparse unmixing, as outlined in \cite{ISTA_Unrol_SUn} and \cite{SU_Unroll}. In these works, an unrolling technique was employed to address the nonnegative $\ell_1$ sparse regression problem, i.e., SUnSAL. To enhance spatial information integration, an intermediate convolutional layer was applied to the abundance representation. The training of the shallow network involved a combination of loss functions, including SAD (Sum of Absolute Differences), MSE (Mean Squared Error), and SID (Spectral Information Divergence).

Furthermore, in \cite{ISTA_Unrol}, a similar unrolling approach was employed, where the Iterative Soft-Thresholding Algorithm (ISTA) \cite{ISTA} was unrolled to tackle the nonnegative $\ell_1$ sparse regression problem. Additionally, in the pursuit of sparse unmixing, the unrolling technique was applied to SUnSAL, as documented in \cite{SUNSAL_Unrol}.

A significant concern arises with the techniques mentioned above when the endmembers do not align with those in the library. In such cases, even a well-curated and pruned spectral library may fall short in representing all the unique endmembers present in real-world datasets. Factors such as noise, atmospheric effects, variations in illumination, and intrinsic material differences introduce shifts in the endmembers. This often leads to scaling discrepancies between the endmembers in the scene and those in the library.

We recently introduced Sparse Unmixing using Archetypal Analysis (SUnAA) to tackle this issue. SUnAA assumes that the endmembers can be expressed as convex combinations of the library endmembers. It addresses the problem as a nonconvex optimization using a cyclic descent algorithm. 
SUnAA runs on the CPU and can be computationally demanding. The number of pixels and endmembers in the dataset influences its performance. In this paper, we propose efficient unmixing algorithms using ADMM to address those drawbacks. Overall, our main contributions can be summarized in three key aspects:
\begin{enumerate}
 \item We introduce a new linear model for semisupervised/library-based unmixing which takes into account the endmember library mismatch and ASC. Our experiments show the advantages of this model compared with the sparse and redundant model used in conventional sparse unmixing.
\item
We propose two ADMM-based methods named Fast Semisupervised Unmixing (FaSUn) and Fast Sparse Unmixing using Soft-Shrinkage (SUnS) aiming at comparing two different priors on the new model. FaSUn and SUnS enforce convexity and sparsity on the endmembers, respectively. Our experiments reveal that the convexity constraint outperforms the sparsity prior.
   \item We provide GPU (PyTorch)-based implementations for the ADMM-based algorithms showcasing the efficiency of the proposed algorithms compared to the state-of-the-art semisupervised unmixing techniques. 

\end{enumerate}

\section{Methodology}
\label{sec: model}

\subsection{Low-rank Linear Mixture Model}
Assuming matrix ${\bf E}\in \mathbb{R}^{p\times r}$ contains $r$ endmembers within the observed hyperspectral pixel ${\bf y} \in  \mathbb{R}^{p}$ (i.e., the sensor has $p$ bands), then linear mixture model (LMM) is given by
 \begin{equation}\label{eq: M0}
{\bf y} = {\bf E}{\bf a} + {\bf n}, ~~{\rm s.t.~~}\sum_{i=1}^ra_i=1, a_i\geq 0, i=1,2,..,r,
\end{equation} 
where ${\bf n}$ denotes the $p$-dimensional random vector denoting the additive random Gaussian noise. To represent all the pixels we use the matrix notation ${\bf Y}$. Then, we have \blue{\cite{SUnAA}}
 \begin{equation}\label{eq: LR}
{\bf Y} = {\bf E}{\bf A} + {\bf N}, ~~{\rm s.t.~~}{\bf A}\geq 0,{\bf 1}_{r}^{T}{\bf A}={\bf 1}_{n}^{T},
\end{equation} 
where {\bf Y}$ \in  \mathbb{R}^{p\times n}$ is the observed HSI, with $n$ pixels and $p$ bands, {\bf N}  $\in \mathbb{R}^{p\times n}$ is noise, and {\bf A} $\in \mathbb{R}^{r\times n}$  contain the $r$ endmembers and their fractional abundances, respectively. ${\bf 1}_n$ indicates an $n$-component column vector of ones. LMM is often used for supervised modeling \cite{HySUPP}. 

\subsection{Sparse and Redundant Linear Mixture Model}
Sparse and Redundant Linear Mixture Model is given by 
 \begin{align}\label{eq: SR}\nonumber
&{\bf Y} = {\bf D}{\bf X} + {\bf N}, \\&
~~~{\rm s.t.}~~~{\bf X}\geq 0,{\bf 1}_{m}^{T}{\bf X}={\bf 1}_{n}^{T},  
\end{align} 
where $ {\bf D}  \in \mathbb{R}^{p\times m} $  ($p\ll m$) denotes the spectral library containing $m$ endmembers and $ {\bf X}\in \mathbb{R}^{m\times n}$ is the unknown  fractional abundances to estimate. Please note that ${\bf D}$ serves as an overcomplete dictionary, and as such, it should be meticulously crafted. A well-structured dictionary comprises endmembers representing the materials present in the scene and can efficiently reduce the redundancy in ${\bf X}$. Consequently, it becomes possible to prune a spectral library based on the spectral angles between spectra, meaning that spectra with small angular differences are removed. However, there is a caveat: this pruning strategy carries the risk of losing endmember materials if they happen to be scaled versions of each other.

In the context of a well-designed dictionary, the pixels in the scene are composed of a mixture of a few dictionary atoms. This characteristic results in ${\bf X}$ being a sparse matrix. It is worth noting that if a specific endmember material is absent from the observed spectra, the corresponding row in ${\bf X}$ will be entirely composed of zeros. This is a common occurrence since abundance values are typically sparse in this model. This framework is frequently employed in the context of sparse unmixing, where fractional abundances ${\bf X}$ are estimated by applying sparsity-enforcing penalties or constraints within a sparse regression formulation.

\subsection{A New Linear Mixture Model}
To enforce both ASC and sparsity, recently, a new linear model inspired by archetypal analysis \cite{RAA} was proposed in \cite{SUnAA} for library-based unmixing (semisupervised). In  \cite{SUnAA}, we proposed a mixing model for observed spectra in
 \begin{equation}\label{eq: M2}
{\bf Y} = {\bf D}{\bf B}{\bf A} + {\bf N}, 
\end{equation} 
 ${\bf B}\in \mathbb{R}^{m\times r}$, determines the contributions of the endmembers from ${\bf D}$. Model (\ref{eq: M2}) exploits both low-rank property and sparse contribution of the endmembers.  The low-rank property of model (\ref{eq: M2}) decreases the computational time. However, the problem will turn into a non-convex problem (both ${\bf B}$ and ${\bf A}$ are unknown), increasing the algorithms' complexity. There are two main advantages of the model (\ref{eq: M2}) compared to the sparse and redundant model: 1- The ASC can be enforced, 2- It can compensate for the mismatch between the library endmembers and data endmembers \cite{SUnAA}. \blue{Incorporating matrix B into the model formulation allows the model to account for the mismatch between data endmembers and library endmembers. The classical sparse and redundant LMM, i.e., ${\bf D}{\bf X}$, assumes that the library endmembers should match the data endmembers. However, in a real application, the actual endmembers that exist in the scene differ from the library ones (endmember variability). Applying different prior on matrix ${\bf B}$ allows different linear combinations of the library endmembers to represent the data endmembers. On the other hand, in model (\ref{eq: M2}) the number of endmembers is unknown and challenging to estimate. In the literature, this problem was addressed with 
 alternative names such as hyperspectral subspace identification, intrinsic
order selection, virtual dimension, and estimation of the number of spectrally distinct
signal sources \cite{HySime,HySURE, Sub_id_Chang_Du}. In practice, this problem can be addressed using eigenvalue-based detection techniques \cite{HFC,Sub_id_Chang_Du,HFC_2} or estimating the mean square errors \cite{HySime, HySURE}. Geometrical-based approaches were also proposed for endmember estimation\cite{ICE,Zare_ICE,GENE}. Indeed, this is an advantage of the conventional sparse and redundant model (\ref{eq: SR}).}

\subsection{FaSUn: Fast Semisupervised Unmixing}
\label{subsec: FaSUn}
We propose a nonconvex optimization to  simultaneously estimate ${\bf B}$ and ${\bf A}$:
\begin{align}\label{eq: M3}\nonumber
  &(\hat{\bf B},\hat{\bf A})=\arg\min_{{\bf B,A}} \frac{1}{2} || {\bf Y}-{\bf DBA}||_{F}^{2} \\&
{\rm s.t.}{\bf B}\geq 0,{\bf 1}_{m}^{T}{\bf B}={\bf 1}_{r}^{T},  {\rm and }{\bf A}\geq 0,{\bf 1}_{r}^{T}{\bf A}={\bf 1}_{n}^{T}.
\end{align}
Note that, in (\ref{eq: M3}), the unknown endmembers are a convex combination of the library's endmembers due to the non-negativity and sum to one constraint on ${\bf B}$. 
In \cite{SUnAA}, we proposed a parameter-free solution to (\ref{eq: M3}) using active set methods. The major issue with SUnAA is that it runs on the CPU and is highly time-consuming. 
Here, we propose an ADMM-based solution for the proposed minimization problem  (\ref{eq: M3}), which benefits from a GPU-accelerated implementation. 

First, we should note that the minimization problem (\ref{eq: M3}) is non-convex; however, it can be solved in two steps using a cyclic descent scheme; ${\bf A}$-step: when ${\bf B}$ is fixed and ${\bf B}$-step: ${\bf A}$ is fixed. In every step, we are dealing with a convex optimization, and therefore, every solution of the steps successively decreases the loss function, which leads to a minimum. The convergence of the final solution is guaranteed upon the convergence of every step throughout the iterations.

${\bf A}$-step: when ${\bf B}$ is fixed then ${\bf E}={\bf DB}$ is fixed. Therefore, problem \ref{eq: M3} turns to
\begin{align}\label{eq: FCLSU}\nonumber
  &\hat{\bf A}=\arg\min_{{\bf A}} \frac{1}{2} || {\bf Y}-{\bf EA}||_{F}^{2} \\&
{\rm s.t.}{\bf A}\geq 0,{\bf 1}_{r}^{T}{\bf A}={\bf 1}_{n}^{T}.
\end{align}
Problem (\ref{eq: FCLSU}) can be solved using any convex optimization, least squares, or quadratic programming solver. However, for unmixing and particularly Earth observation applications, we are often dealing with big datasets, and therefore, these general-purpose convex optimization solvers are not efficient. Here, we propose an ADMM solution. 

To solve problem  (\ref{eq: FCLSU}), we start by splitting ${\bf A}$, 
    \begin{align}\label{eq: FCLSU2}\nonumber
 & \hat{\bf A},\hat \bS=\arg\min_{{\bf A},{\bS}} \frac{1}{2} || {\bf Y}-{\bf E}{\A}||_{F}^{2}\\ &~~ {\rm s.t.}~~ \A=\bS,~~
{\bf S}\geq 0,{\bf 1}_{r}^{T}{\bf A}={\bf 1}_{n}^{T}.
\end{align}
Using ADMM, the augmented Lagrangian (AL) can be written as  
\begin{align}\label{eq: FCLSU3}\nonumber
 & \hat{\bf A},\hat \bS=\arg\min_{{\bf A},{\bS}} \frac{1}{2} || {\bf Y}-{\bf E}{\A}||_{F}^{2}+\frac{\mu}{2} ||\bS-\A-\bL||_{F}^{2}\\ &~~ {\rm s.t.}~~
{\bf S}\geq 0,{\bf 1}_{r}^{T}{\bf A}={\bf 1}_{n}^{T},
\end{align}
where $\bL$ is the Lagrange multiplier. Note that, we did not use AL for ASC. As can be seen in Appendix \ref{app: QuEC}, using Lagrangian and Karush-Kuhn-Tucker (KKT) will lead to a  close form solution and the augmented term in (\ref{eq: FCLSU3}) turns the matrix needs to be inverted non-singular. The solution to this problem is given in three steps.

When \blue{$\bS$} is fixed, the problem turns to 
\begin{align}\label{eq: FCLSU4}\nonumber
 & \hat{\bf A}=\arg\min_{{\bf A}} \frac{1}{2} || {\bf Y}-{\bf E}{\A}||_{F}^{2}+\frac{\mu}{2} ||\bS-\A-\bL||_{F}^{2}\\ &~~ {\rm s.t.}~~{\bf 1}_{r}^{T}{\bf A}={\bf 1}_{n}^{T},
\end{align}
Problem (\ref{eq: FCLSU4}) is a quadratic programming (also known as least squares) with equality constraint (QuEC). In appendix \ref{app: QuEC} we show that there is a closed-form solution for (\ref{eq: FCLSU4}) which is given by 
\begin{equation}\label{eq: QuEC}
  {\bf A}=QuEC(\A,\bS,\bL;\Y,\E,\mu),
\end{equation}
where $QuEC$ is the function given by
\begin{align}\label{eq: QuEC2}\nonumber
 & QuEC(\A,\bS,\bL;\Y,\E,\mu)=\\ &(\Q+\Q\1_r {\textnormal c}\1_r^T\Q)(\E^T\Y+\mu(\bS-\bL)) -\Q\1_r {\textnormal c}\1_n^T,
\end{align}
and
\begin{align}\label{eq: QuEC3}\nonumber
 & \Q=(\E^T\E+\mu\I_r)^{-1}\\ & {\textnormal c}=-1/(\1_r^T\Q\1_r),
\end{align}
When A is fixed, the problem turns to 
\begin{equation}\label{eq: FCLSU6}\nonumber
  \hat \bS=\arg\min_{{\bS}} \frac{\mu}{2} ||\bS-\A-\bL||_{F}^{2}~~ {\rm s.t.}~~
{\bf S}\geq 0,
\end{equation}
and the solution is given by 
\begin{equation}
    \label{eq: FUnmix2}
\bS=\max(0,\A+\bL),
\end{equation}
Finally, we update the multiplier 
\begin{equation}
    \label{eq: L}
\bL=\bL+\A-\bS.
\end{equation}


${\bf B}$-step: When ${\bf A}$ is fixed,      
 problem (\ref{eq: M3}) turns to
    \begin{align}\label{eq: Bstep}\nonumber
  &\hat{\bf B}=\arg\min_{{\bf B}} \frac{1}{2} || {\bf Y}-{\bf DBA}||_{F}^{2}\\ &~~ {\rm s.t.}~~
{\bf B}\geq 0,{\bf 1}_{m}^{T}{\bf B}={\bf 1}_{r}^{T}.
\end{align}
Splitting the variables as $\B=\bS_1$ and $\D\B=\bS_2$, the AL is given by 
    \begin{align}\label{eq: BstepAL}\nonumber
  &\hat{\bf B}=\arg\min_{\B,\bS_1,\bS_2} \frac{1}{2} || {\bf Y}-\bS_2\A||_{F}^{2}+\frac{\mu}{2}|| \bS_1-\B-\bL_1||_{F}^{2}\\ &+\frac{\mu_1}{2}|| \bS_2-\D\B-\bL_2||_{F}^{2}~~ {\rm s.t.}~~
\bS_2\geq 0,{\bf 1}_{m}^{T}{\bf B}={\bf 1}_{r}^{T}.
\end{align}
We solve (\ref{eq: BstepAL}) with respect to each unknown matrix separately. Therefore, we have 
\\
\begin{equation}\label{eq: QuEC4}
  \hat\B=QuEC(\bS_1,\bL_1;(\bS_2-\bL_2),\D,\mu_1/\mu_2),
  \end{equation}
\begin{equation}
    \label{eq: BstepS2}
\hat\bS_1=\max(0,\B+\bL_1),
\end{equation}
\begin{equation}
   \hat \bS_2=(\Y\A^T+\mu_2(\D\B+\bL_2))(\A\A^T+\mu_2\I_r)^{-1}.
\end{equation}
Finally, we update the multipliers
\begin{equation}
    \label{eq: L2}
\bL_1=\bL_1+\B-\bS_1,
\end{equation}
\begin{equation}
    \label{eq: L3}
\bL_2=\bL_2+\D\B-\bS_2.
\end{equation}
Here, we initialize $\bS_1$, $\bS_2$, $\bL_1$, and $\bL_2$ with $0$. $\A$-Step and $\B$-Step should be repeated until the convergence otherwise, the cyclic descent with respect to $\A$ and $\B$ may fail due to the non-convex nature of the problem. The pseudo-code for FaSUn is given in Algorithm \ref{Alg:FaSUn}. Note that for FaSUn the number of endmembers should be given. 

\begin{algorithm}
[tbp]\footnotesize
\SetAlgoLined
\vspace{.cm}
\KwIn{${\bf Y}$: Hyperspectral data, ${\bf D}$: Endmember library, $r$: Number of endmembers, $\mu$, $\mu_1$, and $\mu_2$: AL parameters.}
\vspace{.cm}
\KwOut{$\hat{\bf A}$: Abundances, $\hat{\bf E}$: Endmembers, $\hat{\bf B}$: Endmembers' contributions.}
\textbf{Initialization}: ${\bf B}=0$, ${\bf S}=0$, ${\bf L}=0$, $\bS_i=0$, $\bL_i=0$, $i=1,2$ 

\For{$t = 1$ \KwTo $T$}{
\textbf{A-step} :\\ 
\For{$i=1$ \KwTo $T_1$}{
 $\A=QuEC(\A,\bS,\bL;\Y,\D\B,\mu)$
 \\
$\bS=\max(0,\A+\bL)$\\
$\bL=\bL+\A-\bS$
}
\textbf{B-step} :\\ 
\For{$i=1$ \KwTo $T_2$}{$\B=QuEC(\bS_1,\bL_1;(\bS_2-\bL_2),\D,\mu_1/\mu_2)$\\
$\bS_1=\max(0,\B+\bL_1)$\\
$\bS_2=(\Y\A^T+\mu_2(\D\B+\bL_2))(\A\A^T+\mu_2\I_r)^{-1}$\\
$\bL_1=\bL_1+\B-\bS_1$\\
$\bL_2=\bL_2+\D\B-\bS_2$
}
}
$\hat{\E} = \D \hat{\B}$
\caption{FASUn}
\label{Alg:FaSUn}
\end{algorithm}



\subsection{SUnS: Sparse Unmixing Using Soft-Shrinkage}

Conventional sparse unmixing \cite{SUnSAL} uses model (\ref{eq: SR}) and sparse regression given by 
\begin{align}\label{eq: SUnSAL}\nonumber
  &\hat{\bf X}=\arg\min_{{\bf X}} \frac{1}{2} || {\bf Y}-{\bf DX}||_{F}^{2}+\lambda ||{\bf X}||_1\\&
~~~{\rm s.t.}~~~{\bf X}\geq 0,{\bf 1}_{m}^{T}{\bf X}={\bf 1}_{n}^{T},
\end{align}
to estimate the abundances. An ADMM-based algorithm was proposed to solve the problem (\ref{eq: SUnSAL}), and therefore, it was called sparse unmixing by variable splitting and augmented Lagrangian (SUnSAL). However, it is suggested to use SUnSAL without ASC due to the conflict with $\ell_1$ \cite{JakubLq}.  Additionally, ASC was found to be a rigorous constraint that often does not occur in the real world due to noise and signature variability \cite{SUn}. Therefore, SUnSAL often refers to the problem (\ref{eq: SUnSAL}) without ASC. 
We should note that ignoring ASC breaks physical constraints on pixels for the mixture model. Here, we propose a solution to this challenge. We propose to use an archetypal-type model, i.e., using model (\ref{eq: M2}) and enforce the sparsity on ${\bf B}$ instead of sum to one. In this way, we can keep ASC while enforcing sparsity. \blue{Enforcing $\ell_1$ sparsity on abundance does not allow enforce ASC which is a physical constraint on pixels for the mixture model. ASC does not have to be strict due to noise, however, entirely omitting this constraint may cause abundance estimations that cannot be physically interpreted. Instead, enforcing the sparsity on ${\bf B}$ allows us to apply ASC on ${\bf A}$. Assuming ${\bf E=DB}$, we enforce the sparsity on matrix ${\bf B}$ such that only a few atoms of D contribute to the reconstruction of ${\bf E}$.} Therefore, we propose a new  optimization, 
\begin{align}\label{eq: SUnS1}\nonumber
  &(\hat{\bf B},\hat{\bf A})=\arg\min_{{\bf B,A}} \frac{1}{2} || {\bf Y}-{\bf DBA}||_{F}^{2} +\lambda ||{\bf B}||_1 \\&
{\rm s.t.} {\bf A}\geq 0,{\bf 1}_{r}^{T}{\bf A}={\bf 1}_{n}^{T}, 0\leq{\bf B}\leq 1.
\end{align}
One of the main differences of (\ref{eq: SUnS1}) compared to the conventional sparse unmixing is that the former is nonconvex while the latter is convex if the prior is convex. Moreover, in the former approach, the number of endmembers must be predetermined. 

Here, we propose an ADMM-based solution to (\ref{eq: SUnS1}). Similar to FaSUn, we use a cyclic descent algorithm, and the ${\bf A}$-step is the same and therefore we do not repeat it. For the ${\bf B}$-step, after splitting the variables (${\B}={\bS_1}$ and ${\D\B}={\bS_2}$) the AL is given by
\begin{align}\label{eq: SUnS2}\nonumber
  &\arg\min_{{\bf B,\bS_1,\bS_2}} \frac{1}{2} || {\bf Y}-\bS_2{\bf A}||_{F}^{2} +\lambda ||{\bf S_1}||_1 +\frac{\mu_1}{2} ||{\bS_1}-{\B}-\bL_1||_2^2\\&
+\frac{\mu_2}{2} ||{\bS_2}-{\D\B}-\bL_2||_2^2, {\rm s.t.}  0\leq\bS_1\leq 1.
\end{align}

The solutions w.r.t. each variable are given by  

$\B$-Step:\\
\begin{equation}
  \hat\B=(\mu_1\I_m+\mu_2\D^T\D)^{-1}(\mu_1(\bS_1-\bL_1)+\mu_2\D^T(\bS_2-\bL_2)),
  \end{equation}

\begin{equation}
\hat\bS_1=\min(1,\max(0,soft(\B+\bL_1,\frac{\lambda}{\mu_1}))),
\end{equation}
\begin{equation}
   \hat \bS_2=(\Y\A^T+\mu_2(\D\B+\bL_2))(\A\A^T+\mu_2\I_r)^{-1}.
\end{equation}
\begin{equation}
\bL_1=\bL_1+\B-\bS_1,
\end{equation}
\begin{equation}
\bL_2=\bL_2+\D\B-\bS_2.
\end{equation}

Here, we initialize $\bS_1$, $\bS_2$, $\bL_1$, and $\bL_2$ with $0$. Similarly, $\A$-Step and $\B$-Step should be repeated until the convergence otherwise, the cyclic descent with respect to $\A$ and $\B$ may fail due to the non-convex nature of the problem. The pseudo-code for SUnS is given in Algorithm \ref{Alg:SUnS}. The initializations are the same as Algorithm \ref{Alg:FaSUn}. Note that for SUnS the number of endmembers should be given. 

\begin{algorithm}
[tbp]\footnotesize
\SetAlgoLined
\vspace{.cm}
\KwIn{${\bf Y}$: Hyperspectral data, ${\bf D}$: Endmember library, $r$: Number of endmembers, $\mu$, $\mu_1$, and $\mu_2$: AL parameters.}
\vspace{.cm}
\KwOut{$\hat{\bf A}$: Abundances, $\hat{\bf E}$: Endmembers, ${\bf B}$: Endmembers' contributions.}
\textbf{Initialization}: ${\bf B}=0$, ${\bf S}=0$, ${\bf L}=0$, $\bS_i=0$, $\bL_i=0$, $i=1,2$ 
$\Gamma=(\mu_1\I_m+\mu_2\D^T\D)^{-1}$\\
\For{$t = 1$ \KwTo $T$}{
\textbf{A-step} :\\ 
\For{$i=1$ \KwTo $T_1$}{
 $\A=QuEC(\A,\bS,\bL;\Y,\D\B,\mu)$
 \\
$\bS=\max(0,\A+\bL)$\\
$\bL=\bL+\A-\bS$
}
\textbf{B-step} :\\ 
\For{$i=1$ \KwTo $T_2$}{$\B=\Gamma(\mu_1(\bS_1-\bL_1)+\mu_2\D^T(\bS_2-\bL_2)),$\\
$\bS_1=\min(1,\max(0,soft(\B+\bL_1,\frac{\lambda}{\mu_1}))),$\\
$\bS_2=(\Y\A^T+\mu_2(\D\B+\bL_2))(\A\A^T+\mu_2\I_r)^{-1}$\\
$\bL_1=\bL_1+\B-\bS_1$\\
$\bL_2=\bL_2+\D\B-\bS_2$
}
}
$\hat{\E} = \D \hat{\B}$
\caption{SUnShrink (SUnS)}
\label{Alg:SUnS}
\end{algorithm}


\section{Experimental Results}

We employed a total of \blue{five} datasets, comprising three simulated datasets designed to encompass various mixing scenarios and \blue{two} real-world datasets, \blue{Houston}, and Cuprite, which is a well-documented geological site.
The hyperparameters for the chosen methods were fine-tuned as per Table~\ref{tab:hparams}.
For the simulated datasets, we conducted five independent runs, and the results were then averaged.
The standard deviations are indicated through error bars.
We evaluated the performance of eight semi-supervised unmixing methods, selected as follows: SUnSAL~\cite{SUnSAL}, CLSUnSAL~\cite{CoSUn}, MUA\_SLIC~\cite{RABorsoi2019}, S2WSU~\cite{SZhang2018}, SUnCNN~\cite{SUnCNN}, SUnAA~\cite{SUnAA}, SUnS and FaSUn.
The source code used for running SUnSAL, CLSUnSAL, MUA\_SLIC, S2WSU, and SUnCNN is available in the HySUPP toolbox~\cite{HySUPP} for the sake of reproducibility. \blue{Regarding the parameter tuning procedures, it should be noted that we tuned the most effective parameters. 
Indeed, parameter tuning can be performed for all methods up to some levels. 
The conventional sparse unmixing methods were tuned to perform optimally by tuning their sparsifying regularization parameter. 
For SUnCNN, we used the default number of iterations suggested for different noise levels. 
SUnAA is a parameter free technique. 
FaSUn requires parameter setting for $\mu_1$, $\mu_2$, and $\mu_3$ since those may affect the convergence. 
Therefore, those parameters are set based on the loss function value to ensure the decrease of the cost function over the iterations. 
For SUnS the sparsifying regularization parameter $\lambda$ was also tuned and fixed for the simulated and real datasets.}
Moreover, SUnS and FaSUn were implemented using PyTorch and provided in a dedicated Python package called Fast Semisupervised Unmixing (FUnmix), which is open-source and available at \href{https://github.com/BehnoodRasti/FUnmix}{https://github.com/BehnoodRasti/FUnmix}. 

In terms of quantitative evaluation, we employed the signal-reconstruction-error (SRE) measured in decibels (dB) to assess the estimated abundances, defined by:
\begin{equation} \label{eq:SRE}
    \text{SRE}(\A, \hat{\A})=20\log_{10}\frac{\|{\A}\|_F}{\|{\A}-\hat{\A}\|_F}.
\end{equation}
\blue{SRE was chosen over RMSE due to its logarithmic scale which tends to distinguish the performances better when the reconstruction error is low which happens often in the case of simulated datasets. This is a common choice in library-based unmixing literature \cite{SUn}.}
\begin{table*}[h]
    \centering
        \caption{Hyperparameters in different scenarios}
    \begin{tabular}{c|cc}
    \toprule
     Methods/Data    & Simulated & \blue{Real }\\
    \midrule
      SUnSAL   & SNR-dependent (see \href{https://github.com/ricardoborsoi/MUA\_SparseUnmixing}{source code}) & $\lambda = 0.005$\\
      CLSUnSAL   & SNR-dependent (see \href{https://github.com/ricardoborsoi/MUA\_SparseUnmixing}{source code}) & $\lambda = 0.05, \mu=0.01$\\
      MUA\_SLIC   & SNR-dependent (see \href{https://github.com/ricardoborsoi/MUA\_SparseUnmixing}{source code})  & $\lambda_1 = 0.001, \lambda_2 = 0.01, \beta=10, \text{slic\_size}=200, \text{slic\_reg}=0.01$\\
      S2WSU   & SNR-dependent (see \href{https://github.com/ricardoborsoi/MUA\_SparseUnmixing}{source code}) & $\lambda = 0.001$\\
      SUnCNN   & SNR-dependent (see \href{https://github.com/BehnoodRasti/SUnCNN}{source code} & $\text{niters}=20000$\\
      SUnAA   & default (see \href{https://github.com/inria-thoth/SUnAA}{source code})& default \\
      SUnS   & $T=10000, T_A = T_B = 5, \mu_1 = 50, \mu_2 = 2, \mu_3 = 1, \lambda=0.01$ & $T=10000, T_A = T_B = 5, \mu_1 = 400, \mu_2 = 100, \mu_3 = 1, \lambda=0.1$\\
      FaSUn   & $T=10000, T_A = T_B = 5, \mu_1 = 50, \mu_2 = 2, \mu_3 = 1$ & $T=10000, T_A = T_B = 5, \mu_1 = 400, \mu_2 = 20, \mu_3 = 1$ \\
    \bottomrule
    \end{tabular}
    \label{tab:hparams}
\end{table*}

\subsection{Data Description}

\subsubsection{Synthetic Datasets with Spatial Structure}

We simulated two data cubes (DC1 and DC2). 
DC1 was simulated using a linear mixing model with 5 endmembers selected from the USGS library and 75$\times$75 pixels.
The abundance maps are composed of five rows of square regions uniformly distributed over the spatial dimension. 
This dataset contains pure pixels for all endmembers.
\blue{DC1 corresponds to the simplest setting where traditional sparse unmixing methods tend to shine.}
DC2 has 100$\times$100 pixels and was simulated using a linear mixing model with 9 endmembers. 
The abundance maps were sampled from a Dirichlet distribution centered at a Gaussian random field to have piece-wise smooth maps with steep transitions. 
Therefore, DC2 contains spectral variations. 
\blue{DC2 enables us to study the effect of spectral variability in a controlled setting in which the ground truth is available and quantitative results can be obtained.}
For DC1 and DC2, an endmember library $\D \in \mathbb{R}^{224\times 240}$, composed of 240 spectral signatures was selected from the USGS library with a minimum pair-spectra angle of 4.44\textdegree.
Synthetic Gaussian noise is added so as to create different signal-to-noise ratio (SNR) scenarios (\emph{e.g.}, 20, 30, and 40 dB SNR).

\subsubsection{Synthetic Datasets with varying Pixel Purity Levels}

We assessed the performance of our chosen methods in an alternative unmixing scenario characterized by the absence of spatial structure but parameterized pixel purity levels.
In this context, we were able to explore a spectrum of scenarios ranging from highly mixed, where pure pixels are typically missing, to predominantly pure pixels, with various degrees of mixing in between.
The degree of pixel purity was quantified by the parameter $\rho$, with lower values signifying less purity and higher values indicating greater purity.

To construct our dataset, we selected six spectra from the USGS library ($\D \in \mathbb{R}^{224 \times 498}$) and created a dataset of size $n=100\times100$ pixels using the following methodology.
Initially, we generated a substantial number of abundance samples denoted as $\mathcal{S}$.
These samples were drawn from the symmetric Dirichlet distribution, employing a scalar concentration parameter $\alpha = 1 / r$ where $r$ corresponds to the number of endmembers (in this case, 6).
Given a pixel purity level $\rho$, we randomly drew $n$ abundances from $\mathcal{S}$ such that their $\ell_2$ norm fell within the range of $\rho - 0.1$ to $\rho$.
Subsequently, we combined the selected spectra based on the sampled abundances to create the final pixel set $\Y = \left[\yy_1, \ldots, \yy_n \right]$.
Synthetic Gaussian noise is eventually added to obtain an input SNR of 30 dB.
\blue{The objective is to quantify the performance of the proposed methods in a setup where the degree of pixel purity is entirely parametrized. It is used to demonstrate the robustness of our methods in unmixing scenarios that are increasingly harder as $\rho$ decreases.}

\subsubsection{\blue{Real Dataset with Endmember Bundles}}

\blue{The Houston dataset used in this paper is a 105$\times$128 pixels subset of an image acquired over the University of Houston campus, Texas, USA, in June 2012.
The preprocessed hyperspectral image consists of 144 spectral bands in the 380nm to 1050nm region \cite{IADF2013}.
\blue{Houston is challenging owing to the spectral variability of the macroscopic materials present in the scene.
Therefore the spectral library has been designed in a way to provide several dictionary elements per class.}
However, there is no library of reference endmembers associated with this image so we use the procedure proposed by \cite{End_bundle} to automatically extract endmember bundles from the image pixels.
We consider the following four macroscopic materials: concrete, asphalt, vegetation, and red (metallic) roof.
In a nutshell, we leverage VCA \cite{VCA} on ten subsets of the data pixels to estimate four endmembers per subset.
We then apply the k-means algorithm (with the cosine similarity measure) using four classes to cluster the estimated endmembers into four endmember bundles.
The resulting dictionary consists in 40 endmembers, each associated to a cluster, or bundle.}

\subsubsection{\blue{Real Dataset with a Spectral Library}}
The Cuprite dataset used in this paper contains 250$\times$191 pixels. Cuprite is a well-studied mineral site, and dominant minerals are demonstrated using a geological ground reference.
Therefore, the abundance maps estimated by different techniques can be compared visually.
We use a library $\D \in \mathbb{R}^{188\times 498}$ composed of 498 spectral pixels from the USGS library. Note that we remove the water absorption and noisy bands, such that the final pixels are of dimension $p=188$.
\blue{Cuprite has been selected on account of its challenging characteristics. With almost 500 dictionary elements (from the USGS library), conventional sparse unmixing methods tend to underestimate the abundance of the predominant materials which are Alunite, Chalcedony, and Kaolinite.
In addition, due to its high number of pixels (more than 47k), Cuprite offers a testing ground for assessing the efficiency of the proposed algorithms.}

\subsection{Experimental Results: Synthetic Datasets}

We first compare the selected methods on the synthetic datasets.
Figure~\ref{fig:DC} summarizes the results in terms of SRE on DC1 and DC2 for different input SNR.
The following observations can be formulated:
\begin{figure*}[h]
  \centering
  \subfloat[DC1]{\includegraphics[width=.9\textwidth]{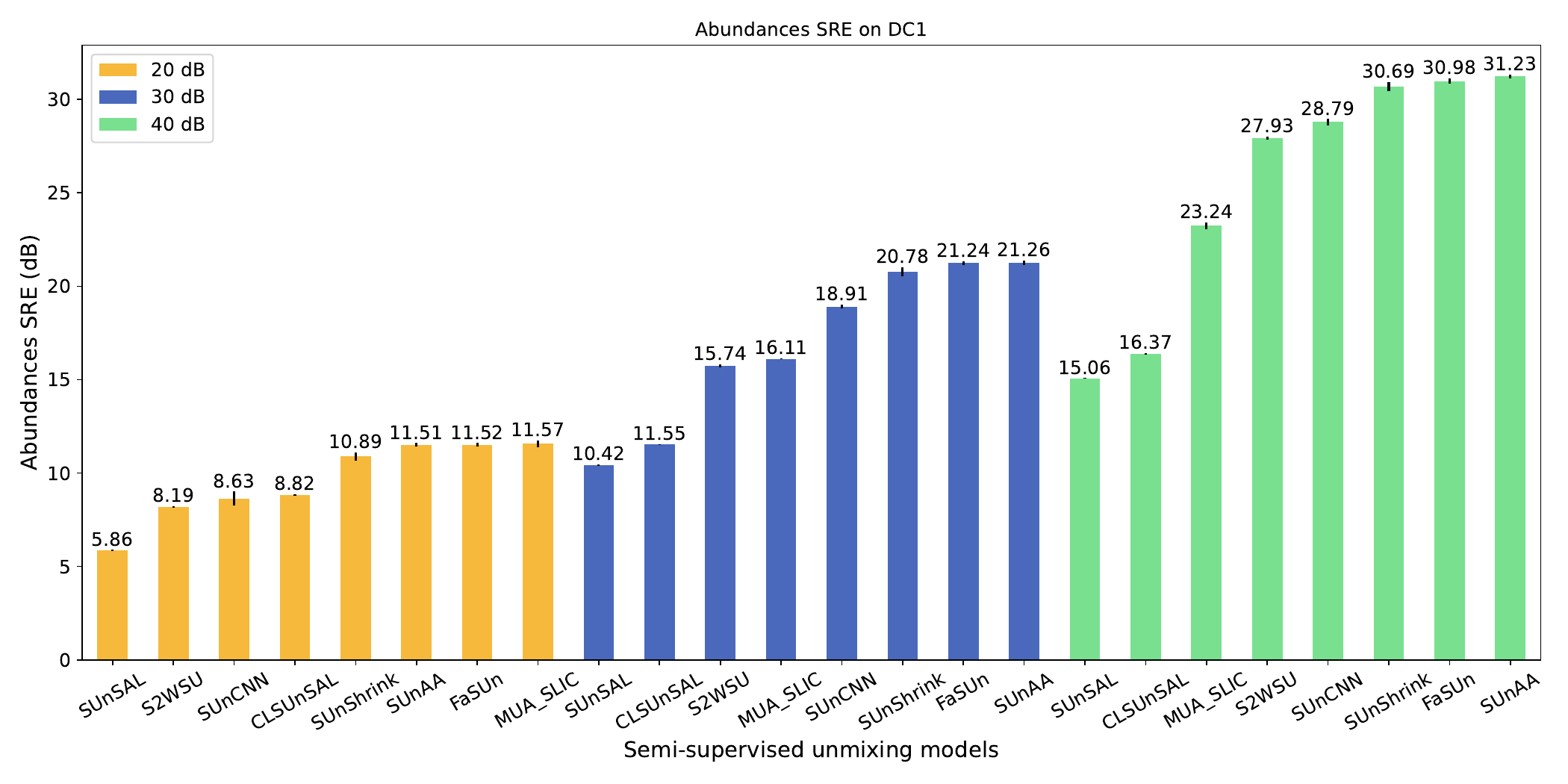}}
  \hfil
  \subfloat[DC2]{\includegraphics[width=.9\textwidth]{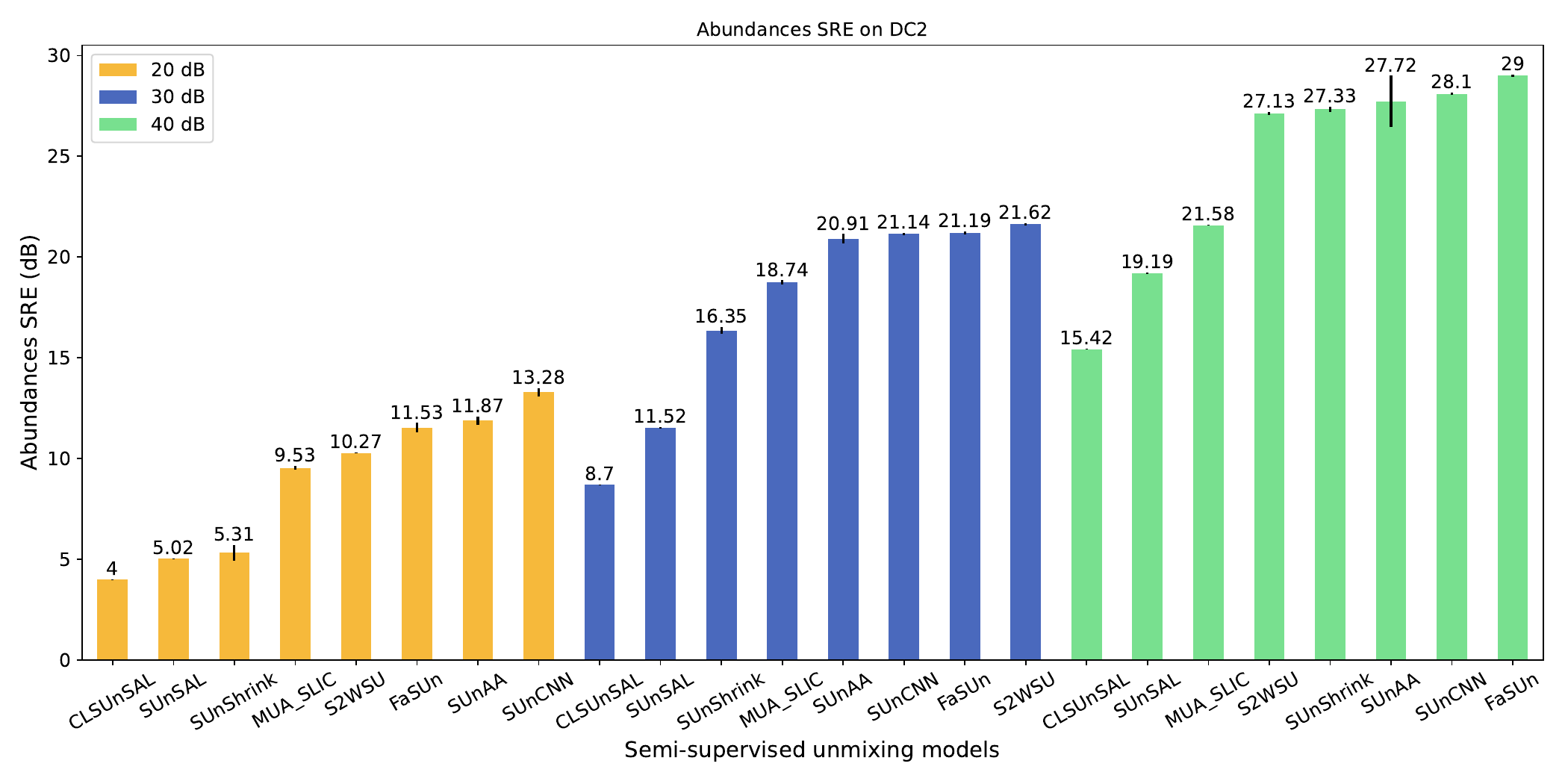}}
  \caption{Abundance SRE ($\uparrow$) in dB for the selected semi-supervised methods on two simulated datasets (DC1 and DC2) for three noise levels.}
  \label{fig:DC}
\end{figure*}

\begin{itemize}
    \item In the presence of pure pixels (Fig.~\ref{fig:DC}, DC1), regardless of the noise level, archetypal analysis inspired techniques (\ie SUnAA, FaSUn, and SUnS) perform very well. 
    MUA\_SLIC is suitable when the noise is significant (\eg 20 dB) but its performance drops when dealing with less noisy images due to its segmentation approach.
    CLSUnSAL performs a bit better than SUnSAL but struggles to compete with the top performing methods overall.
    S2WSU and SUnCNN provide similar results on this dataset.
    It is worth mentioning that most methods, except the archetypal analysis inspired ones, require tuning their regularization parameter to obtain competitive results depending on the input SNR, which is a major hindrance.

    \item In the presence of spectral variability (Fig.~\ref{fig:DC}, DC2), SUnCNN performs very well, likely due to its convolutional architecture that is suited to capture the spectral variability relying on the spatial structure of the data.
    Similarly, S2WSU obtains competitive results.
    SUnAA and FaSUn, which both solve problem (\ref{eq: M3}), outperform the other methods.
    It should be noted that SUnS struggles in the lower SNR scenarios (\ie 20 and 30 dB) which reveals the advantage of convexity constraint compared to the sparsity prior. 
    Moreover, it appears that SUnSAL and CLSUnSAL demonstrate the poorest performances compared to the other methods.
    
\end{itemize}

\begin{figure*}[h]
\centering
\includegraphics[width=.9\textwidth]{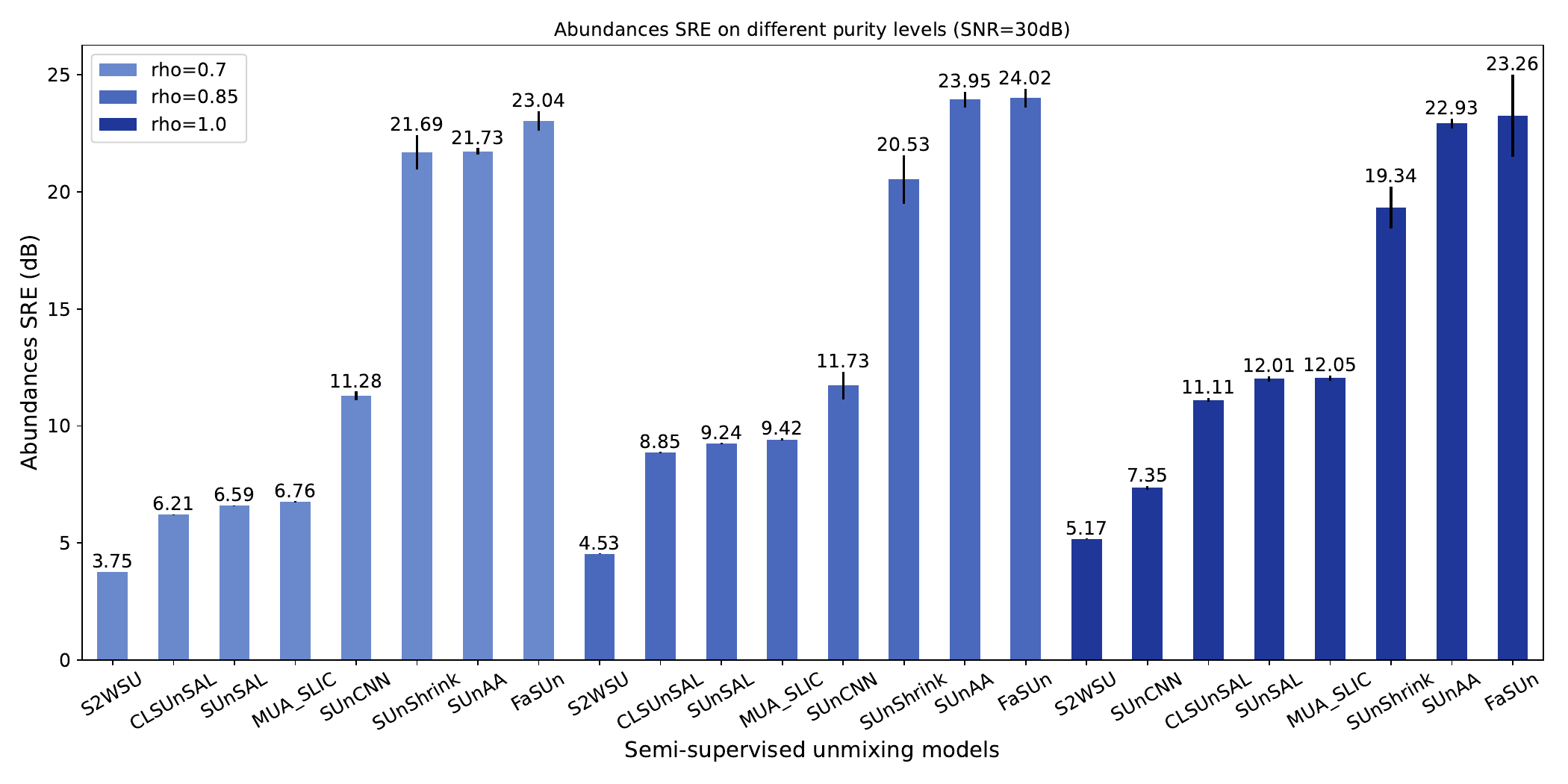}
\caption{Abundance SRE ($\uparrow$) in dB for the selected semi-supervised methods on three different pixel purity levels ($\rho$) for a given input SNR (30 dB)}
\label{fig:MR}
\end{figure*}

Figure~\ref{fig:MR} summarizes the results in terms of SRE for different pixel purity levels using a fixed SNR (30 dB).
The following observations can be made:
\begin{itemize}
    \item The archetypal analysis inspired models severely outperform their sparse unmixing counterparts, regardless of the pixel purity level.
    This is particularly striking when the pixel purity is low, meaning the image only contains highly mixed pixels.
    This indicates that SUnAA, FaSUn and SUnS are better suited to handle highly mixed scenarios.
    Furthermore, the latter methods do not leverage spatial information contrarily to S2WSU, whose performance drops significantly due to the absence of spatial structure in the data.
    
    \item It is worth mentioning that the endmembers library, $\D$, has not been pruned, contrarily to the previous setups, in which the number of atoms in the dictionary went down from $498$ to $240$.
    Therefore there is a clear benefit in having access to the number of endmembers present in the scene (\ie $r$), which is available for SUnAA, FaSUn and SUnShrink.

\end{itemize}

\begin{figure*}[h]
\centering
\includegraphics[width=.8\textwidth]{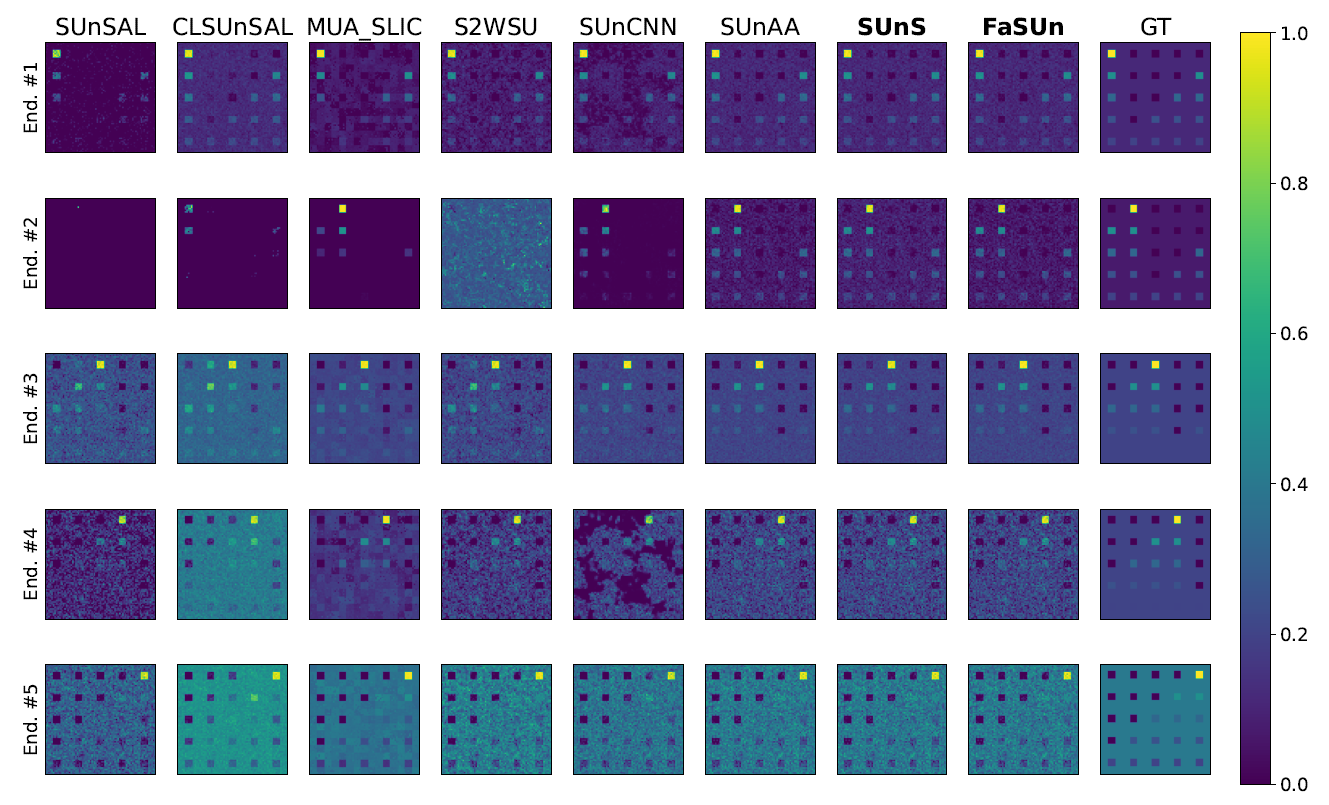}
\caption{Visual comparisons of abundance maps estimated by using different semi-supervised unmixing methods applied to DC1 (20 dB).}
\label{fig:DC1_ab}
\end{figure*}

\begin{figure*}[h]
\centering
\includegraphics[width=.8\textwidth]{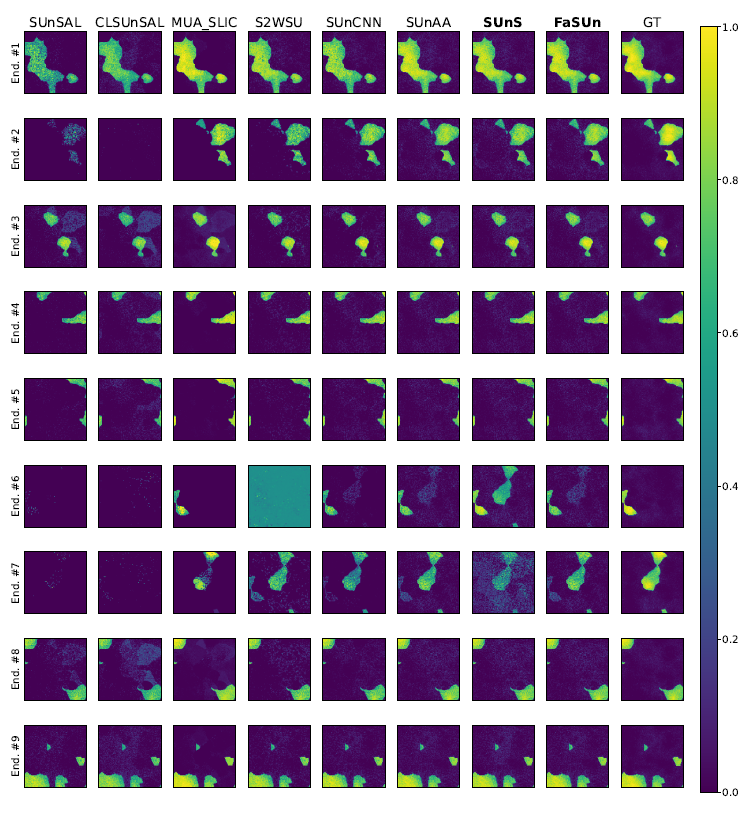}
\caption{Visual comparisons of abundance maps estimated by using different semi-supervised unmixing methods applied to DC2 (20 dB).}
\label{fig:DC2_ab}
\end{figure*}

Figure~\ref{fig:DC1_ab} and \ref{fig:DC2_ab} demonstrate the visual comparisons of estimated abundances by applying different semisupervised unmixing techniques to DC1 and DC2, respectively, for SNR=20 dB. Overall, the visual comparisons confirm that FaSUn and SUnA perform similarly and outperform the other semisupervised techniques for those datasets. SUnS performs similarly to SUnAA and FaSun in the case of DC1. However, in the case of DC2, SUnS cannot successfully estimate the abundance map associated with endmember 7. It is worth mentioning that, in the case of DC1 (20 dB), MUA\_SLIC provides the highest SRE, however, the visual comparisons reveal that abundances are oversmoothed due to the prior segmentation step which can be associated with the high SRE only for low SNR. SUnCNN performs well in the case of DC2 but the abundances for DC1 are oversmooothed. The abundances estimated by SUnSAL, CLSUnSAL, and S2WSU are not competitive with the other methods. 

\subsection{Experimental Results: Real Data}
\begin{figure*}[h]
    \centering
    \includegraphics[width=\textwidth]{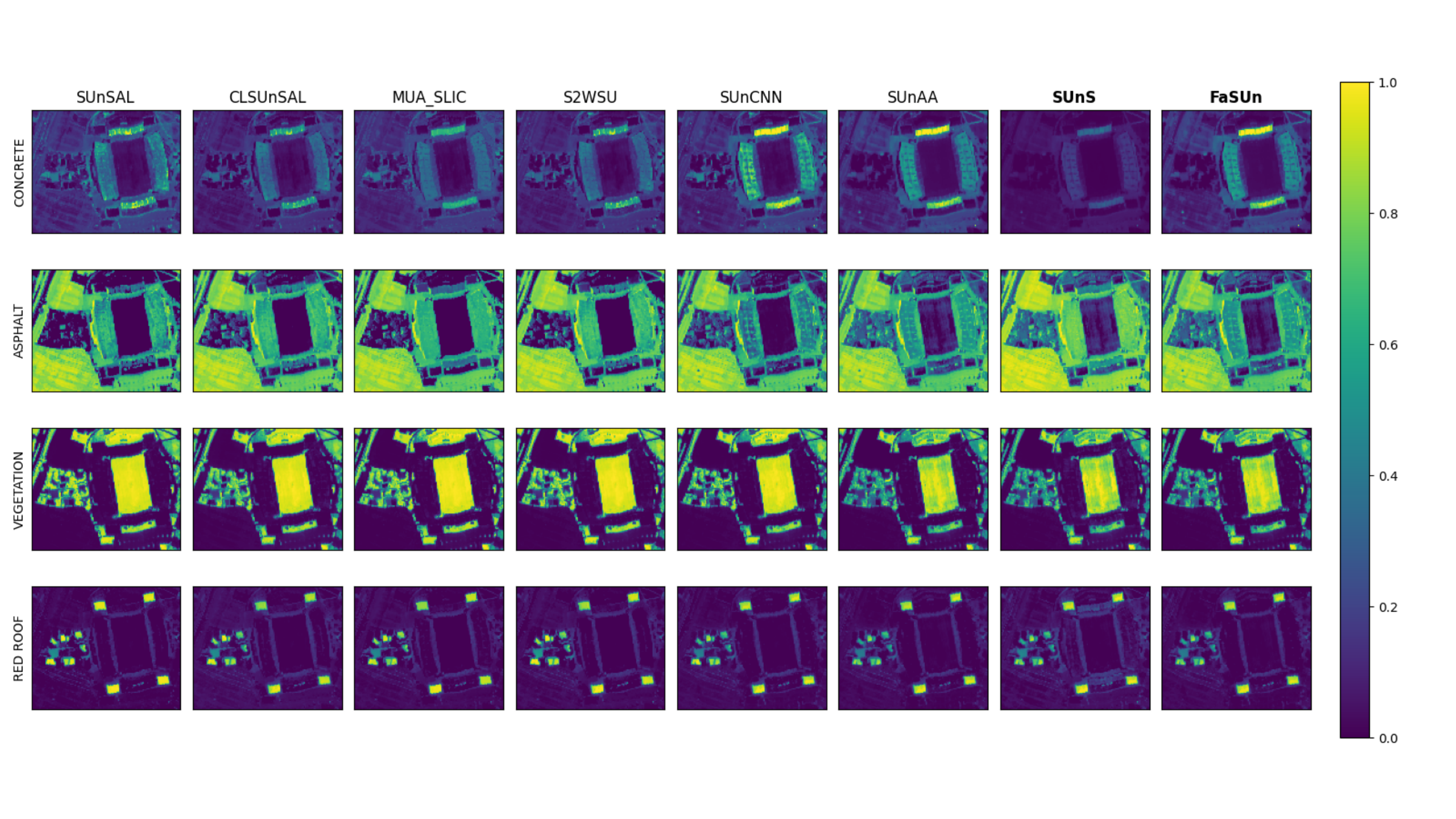}
    \caption{\blue{Estimated abundances obtained by different semi-supervised methods on the Houston dataset.}}
    \label{fig:Houston}
\end{figure*}

\blue{
Figure~\ref{fig:Houston} compares the estimated abundances by applying different semisupervised unmixing methods to the Houston dataset. In this experiment, we use endmember bundles as a library. The abundances are shown by applying a sum to one of the associated abundances to endmember bundles except for SUnAA, SUnS, and FaSUn since they use ASC. The results of SUnAA and FaSUn are similar since they solve the same problem. FaSUn, SUnAA, and SUnCNN demonstrate a higher abundance ratio for Concrete compared to the other techniques. SUnS and SUnSAL provide higher abundance ratios for Asphalt and Red Roof, respectively. The sparse unmixing techniques using the SR model provide a higher abundance ratio for vegetation. Overall, FaSUn SUnAA, and SUnCNN show better visual performance in distinguishing the materials.  
}

\begin{figure*}[h]
\centering
   \subfloat[Reference Map]{\includegraphics[width=.16\textwidth]{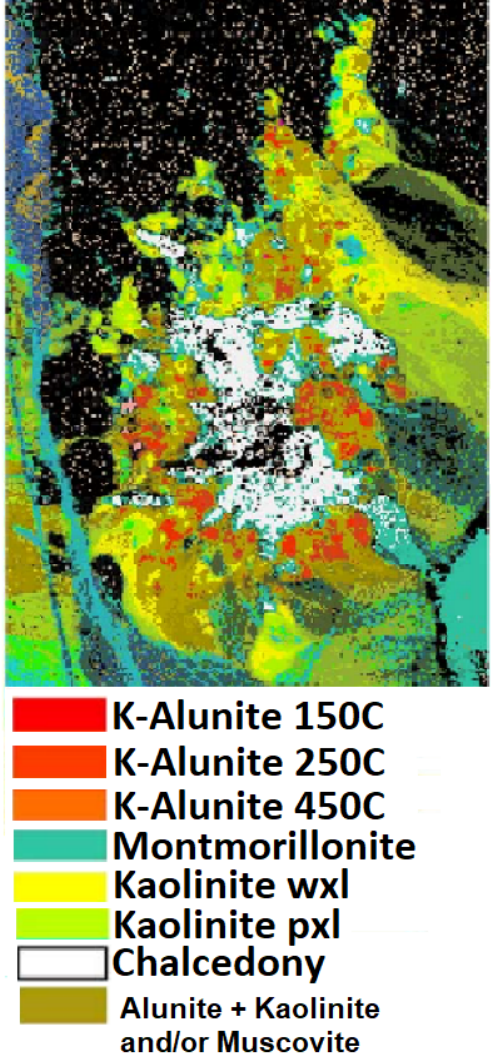}}
   \subfloat[Estimated abundances]{\includegraphics[width=1\textwidth]{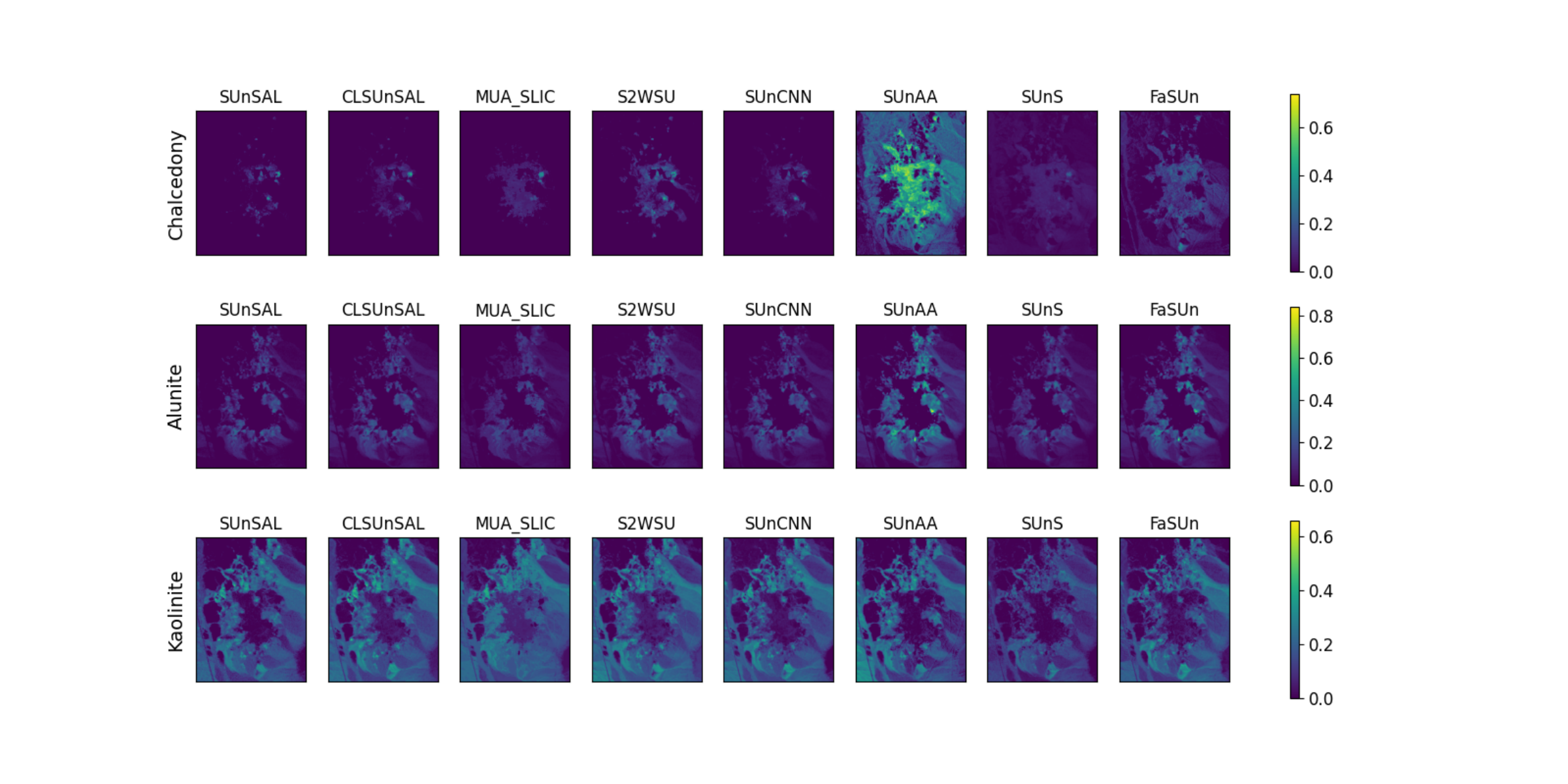}}
  \caption{Estimated abundances obtained by applying different semi-supervised methods to Cuprite compared with the geological reference map.}
\label{fig:Cuprite}
\end{figure*}

Figure~\ref{fig:Cuprite} visually compares the estimated abundances for three dominant materials, \ie, Chalcedony, Alunite and Kaolinite, using the geological map as a reference.
It is worth mentioning that hyperparameters for each method had to be tuned, except for SUnAA as it is parameter-free.
Moreover, for the archetypal analysis-inspired methods to work, the number of endmembers in the scene (\ie $r$) was set to $r=14$. 

Visual comparison based on the reference map reveals that SUnAA better estimated  Chalcedony compared to the other methods.
FaSUn does not exhibit the same saliency as SUnAA, but still detects Chalcedony on a bigger area of the map than other sparse unmixing methods.
As for Alunite, SUnAA and FaSUn both show strong responses to the mineral in the expected areas.
Sharp abundance maps are obtained for Kaolinite by SUnAA and FaSUn which are in line with the reference map.
Overall, SUnS shows similar performances as the other sparse unmixing techniques. It is worth mentioning that SUnS introduces another regularization parameter, $\lambda$, (similar to the other sparse unmixing methods) that requires additional tuning, compared to FaSUn.

\subsection{Processing Time \blue{and Computational Complexity}}

Perhaps the gist of our contributions is the considerable scalability of our proposed approaches (FaSUn and SUnShrink), as highlighted in Table~\ref{tab:timer}. 
Note that SUnCNN processing time depends on the number of iterations, which is itself dependent on the input SNR.
Here we report the processing time for a fixed SNR equal to 30 dB obtained using a computer with an Intel(R) Xeon(R) Silver 4110 CPU at 2.10GHz, 32 cores, 64 Gb of RAM, and a NVIDIA GeForce RTX 2080 Ti GPU with 12 Gb of RAM. 
We observe that SUnAA processing time is prohibitive for real-world applications, as soon as the number of pixels exceeds $n=10000$.
On the contrary, the most efficient methods are SUnS and FaSUn on Cuprite, which exploit GPU computations despite using many outer iterations, \ie $T=10000$. It is worth highlighting the growth rate of consumed time by the FaSUn and SUnS compared to the other techniques. They take around three minutes to perform on a dataset with 90k pixels and 224 bands. These results shed light on the scalability challenge posed to semi-supervised methods despite ongoing efforts. \blue{We should note that among the methods mentioned above, SUnCNN, FaSUn, and SUnS use GPU and the rest use CPU for computations.
Concerning the computational complexity, each iteration of FaSUn and SUnS algorithms is $\mathcal{O}\left((T_1 + T_2) n p r \right)$ due to costly matrix products and solving linear systems.}

\begin{table*}[h]
\centering
\caption{Processing time. The best results are in boldface and the second best are underlined.}
\resizebox{\textwidth}{!}{
\begin{tabular}{c|c|c|c|c|cccccccc}
\toprule
             & \# Pixels & \# Bands & \# Endmembers & \# Atoms & SUnSAL & CLSUnSAL & MUA\_SLIC & S2WSU & SUnCNN & SUnAA & SUnS & FaSUn \\
\midrule
DC1          &      5625     &   224       &      5        &    240    &    \underline{7.1}      &     15.3      &      \textbf{6.1}    &     41.1   &   94.7    &   131.1   & 85.3 & 89.2      \\
DC2          &      10000     &    224      &      9         &    240    &    \underline{18.9}      &      27.2     &    \textbf{10.2}      &    51.0    &   78.0    &   264.8   &   80.0  &  77.6 \\
Mixed Pixels 10k &     10000      &    224      &       6        &    498    &   \textbf{46.4}       &      159.1     &     98.2     &    \underline{61.3}    &   87.2    &   208.7   &    136.9  & 134.3 \\
\blue{Houston} & \blue{13440} & \blue{144} & \blue{4} & \blue{40} & \blue{\underline{10.9}} & \blue{46.4} & \blue{\textbf{5.1}} & \blue{28.4} & \blue{220.4} & \blue{109.6} & \blue{61.8} & \blue{63.1} \\ 
Cuprite      &     47750      &     188     &      14         &    498    &   200.3       &    343.8       &    198.5      &   662.0     &   769.6    &   1838.0   &   \underline{154.5} & \textbf{147.7}   \\

Mixed Pixels 90k &     90000      &    224      &       6        &    498    
&  566.5 & 2065.3 & 755.9 &  660.0  &  576.0   &  1359.0   &    \textbf{183.4}  & \underline{194.6} \\

\bottomrule
\end{tabular}
}
\label{tab:timer}
\end{table*}

\section{Conclusion}
We proposed two nonconvex optimizations for hyperspectral unmixing relying on a new type of linear model called FaSUn and SUnS. FaSun utilizes convexity constraint while SUnS uses a sparse prior. We derived ADMM-based solutions for those problems implemented using PyTorch. We conducted a comprehensive evaluation of our proposed techniques by comparing them with SOTA methods. This evaluation was performed on three simulated datasets, considering spatial structure, spectral variability, and various pixel purity and noise levels, and we measured their performance in terms of signal-to-reconstruction error. The results strongly support the superiority of the proposed unmixing technique, FaSUn, over existing state-of-the-art methods. Notably, FaSUn consistently achieved better performance across a range of scenarios. The results confirmed the advantage of the convexity constraint compared to the sparsity-promoting prior. Additionally, we applied these unmixing techniques to real-world data using the Cuprite dataset. To validate the accuracy and practical utility of the proposed methods, we visually compared the results with geological reference maps. Furthermore, the proposed ADMM-based algorithms demonstrated remarkable efficiency in addition to their superior performance. This was evident when comparing the processing times for large datasets, highlighting the practical advantages of our approach.

 \appendices
\section{Derivation of QuEC Function}
\label{app: QuEC}
 Assuming the quadratic programming (or least
squares) with the equality constraint as 
\begin{align}\label{eq: QuEC1}\nonumber
 & \hat{\bf A}=\arg\min_{{\bf A}} \frac{1}{2} || {\bf Y}-{\bf E}{\A}||_{F}^{2}+\frac{\mu}{2} ||\bS-\A-\bL||_{F}^{2}\\ &~~ {\rm s.t.}~~{\bf 1}_{r}^{T}{\bf A}={\bf 1}_{n}^{T},
\end{align}
the Lagrangian function is given by
\begin{equation}\label{eq: QuEC11}
 \Lagr(\A,\nu)=\frac{1}{2} || {\bf Y}-{\bf E}{\A}||_{F}^{2}+\frac{\mu}{2} ||\bS-\A-\bL||_{F}^{2}+\nu^T({\bf 1}_{r}^{T}{\bf A}-{\bf 1}_{n}^{T}),
\end{equation}
where the solution can be given using Karush-Kuhn-Tucker (KKT) conditions. Here, we derive the KKT conditions for (\ref{eq: QuEC11}). The stationary condition is give by $\nabla_\A \Lagr(\A,\nu)=0$,
\begin{equation}\label{eq: QuEC22}
 (\E^T\E+\mu\I)\A+\1_r\nu=\E^T\Y+\mu(S-L)
\end{equation}
For the primal feasibility we hold ${\bf 1}_{r}^{T}{\bf A}={\bf 1}_{n}^{T}$. Therefore, we have
\begin{equation}
    \begin{pmatrix}
    \E^T\E+\mu\I&\1_r\\
    \1_r^T&0
    \end{pmatrix}
        \begin{pmatrix}
    \A\\
    \nu
    \end{pmatrix}
    =
        \begin{pmatrix}
    \E^T\Y+\mu(\bS-\bL)\\
    \1_n^T
    \end{pmatrix},
\end{equation}
using the blockwise inversion, the solution is given by
\begin{equation}\label{eq: QuEC33}
 \hat\A=(\Q+\Q\1_r {\textnormal c}\1_r^T\Q)(\E^T\Y+\mu(\bS-\bL)) -\Q\1_r {\textnormal c}\1_n^T
\end{equation}
Where
\begin{align}\label{eq: QuEC344}
 & \Q=(\E^T\E+\mu\I_r)^{-1}\\ & {\textnormal c}=-1/(\1_r^T\Q\1_r).
\end{align}
As can be seen, the augmented term in (\ref{eq: QuEC11}), makes matrix $\Q$ to be always non-singular (note that $\mu>0$) and therefore the closed form solution (\ref{eq: QuEC33}) becomes feasible.

\section*{Acknowledgment}
AZ, JM, and JC  were supported by ANR 3IA MIAI@Grenoble Alpes (ANR-19-P3IA-0003). AZ and JM were supported by the ERC grant number 101087696 (APHELAIA project). \blue{We would like to thank the Hyperspectral Image Analysis Laboratory at the University of Houston, and the IEEE
GRSS Image Analysis and Data Fusion Technical Committee for providing the Houston dataset.}

\ifCLASSOPTIONcaptionsoff
  \newpage
\fi



%
\bibliographystyle{IEEEbib}
\bibliography{New, refs,new_ref,refs1,HyDe,refs_misicnet,Ref_non}

%






\vspace{11pt}

\vfill
\begin{IEEEbiography}[{\includegraphics[width=1in,height=1.25in,clip,keepaspectratio]{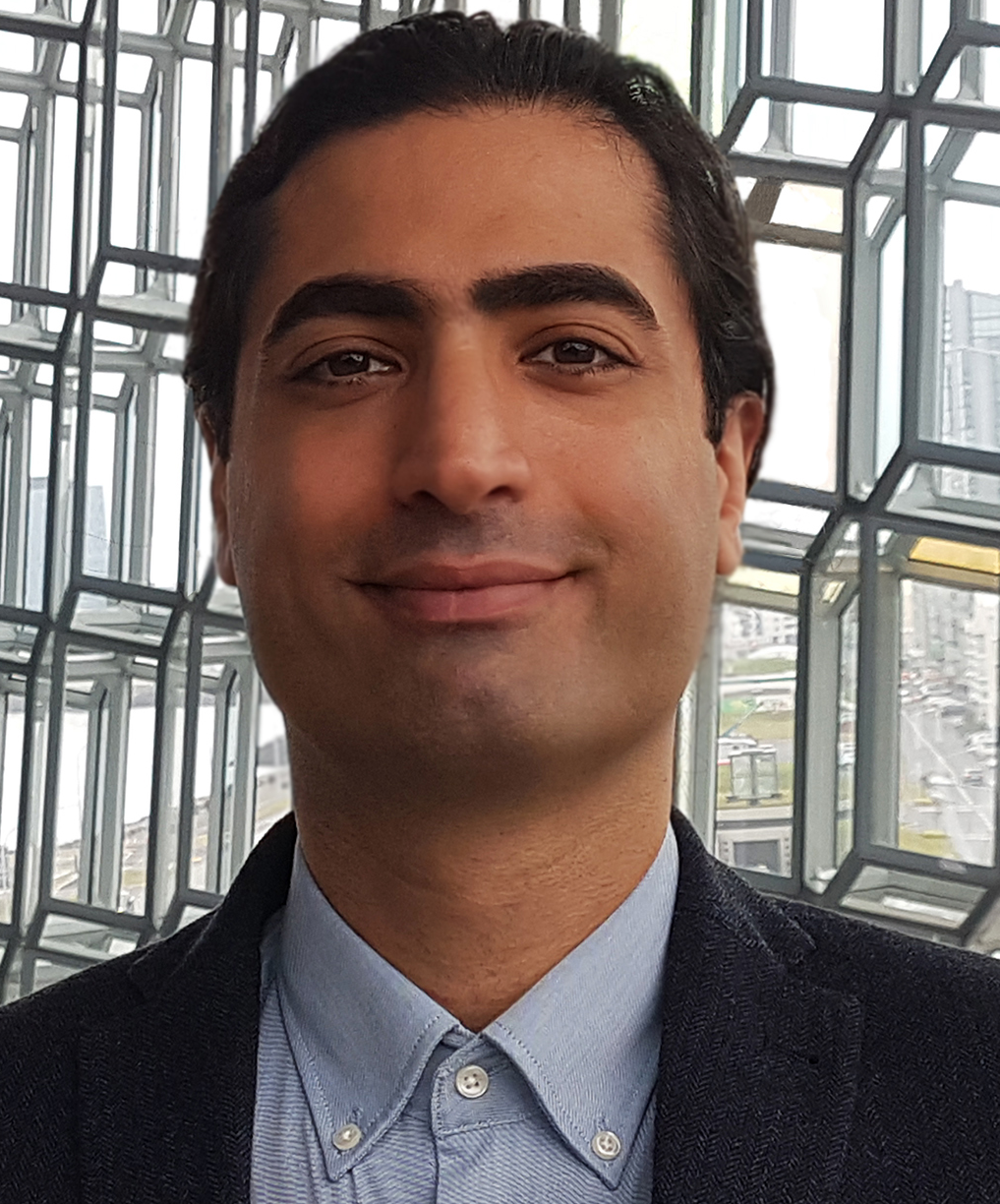}}]{Behnood Rasti (M’12–SM’19)} received the B.Sc. and M.Sc. degrees in electronics and electrical engineering from the Electrical Engineering Department, University of Guilan, Rasht, Iran, in 2006 and 2009, respectively, and the Ph.D. degree in electrical and computer engineering from the University of Iceland, Reykjavik, Iceland, in 2014. He was a Valedictorian as an M.Sc. Student in 2009. In 2015 and 2016, he worked as a Post-Doctoral Researcher with the Electrical and Computer Engineering Department, University of Iceland. From 2016 to 2019, he was a Lecturer with the Center of Engineering Technology and Applied Sciences, Department of Electrical and Computer Engineering, University of Iceland. Dr. Rasti was a Humboldt Research Fellow in 2020 and 2021. From 2022 to 2023, he was a Principal Research Associate with Helmholtz-Zentrum Dresden-Rossendorf (HZDR), Dresden, Germany. He is currently a senior research scientist at the Faculty of Electrical Engineering and Computer Science, Technische Universität Berlin, and the Berlin Institute for the Foundations of Learning and Data, Berlin, Germany. His research interests include signal and image processing, machine/deep learning, remote sensing, and artificial intelligence. 

Dr. Rasti won the Doctoral Grant of the University of Iceland Research Fund “The Eimskip University Fund” and the “Alexander von Humboldt Research Fellowship Grant” in 2013 and 2019, respectively. He serves as an Associate Editor for the IEEE GEOSCIENCE AND REMOTE SENSING LETTERS (GRSL). 
\end{IEEEbiography}

\begin{IEEEbiography}
[{\includegraphics[width=1in,height=1.25in,clip,keepaspectratio]{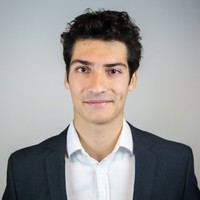}}]{Alexandre Zouaoui} received his M.Sc. degree in computer science from Télécom Paris, Paris, France, in 2019. He received his Ph.D. in applied mathematics at Université Grenoble Alpes, Grenoble, France, in 2024 on Sparse and Archetypal Decomposition Algorithms for Hyperspectral Images Restoration and Spectral Unmixing. His research interests include computer vision, interpretable machine learning and remote sensing image processing. In 2024, he joined Data Science Experts as a research engineer focusing on applications related to hyperspectral imagery.
\end{IEEEbiography}

\begin{IEEEbiography}[{\includegraphics[width=1in,height=1.25in,clip,keepaspectratio]{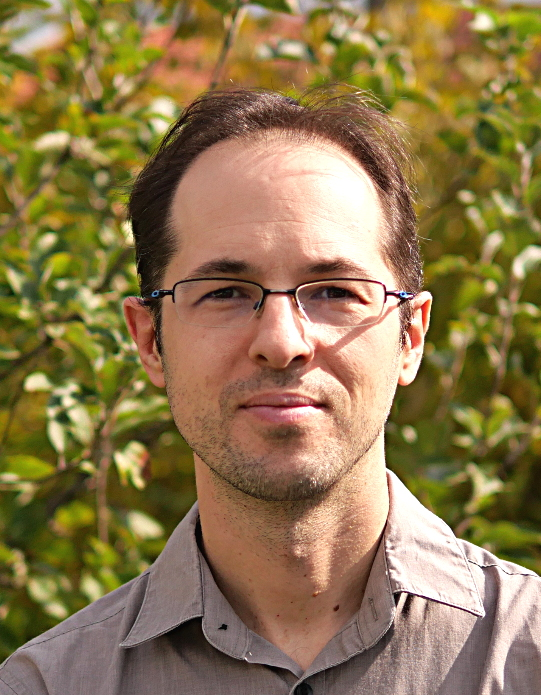}}]{Julien Mairal} received the graduate degree from Ecole Polytechnique, Palaiseau, France, in 2005, and the Ph.D. degree from Ecole Normale Superieure, Cachan, France, in 2010. After that, he joined the statistics department at UC Berkeley as a post-doctoral researcher. In 2012, he joined Inria, Grenoble, France, where he is currently a research director and head of the Thoth team. His research interests include machine learning, computer vision, mathematical optimization, and statistical image and signal processing. He received a starting grant and a consolidator grant from the European Research Council, respectively in 2016 and 2022. He was awarded the Cor Baayen prize in 2013, the IEEE PAMI young research award in 2017 and the test-of-time award at ICML 2019.
\end{IEEEbiography}

\begin{IEEEbiography}[{\includegraphics[width=1in,height=1.25in,clip,keepaspectratio]{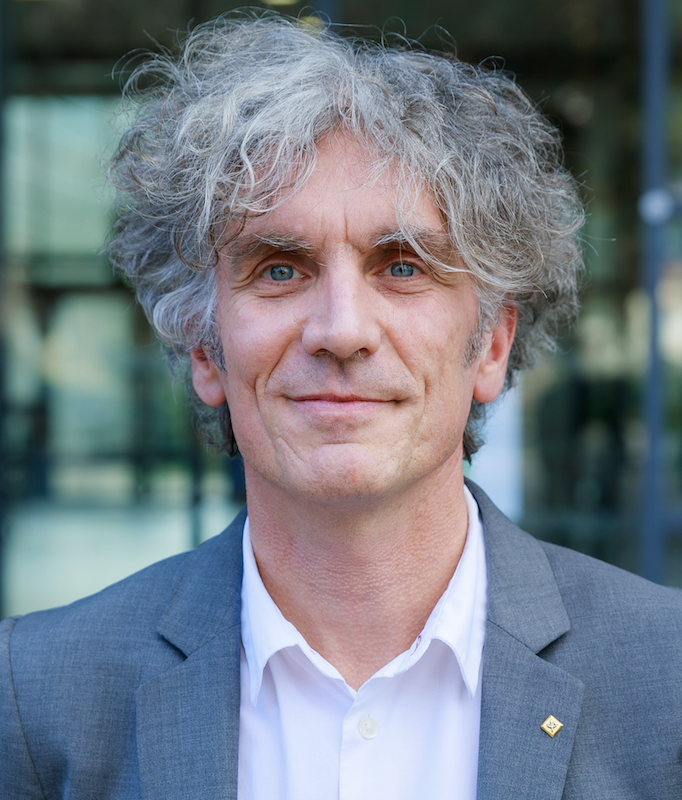}}]{Jocelyn Chanussot (M’04–SM’04–F’12)} received the M.Sc. degree in electrical engineering from the Grenoble Institute of Technology (Grenoble INP), Grenoble, France, in 1995, and the Ph.D. degree from the Université de Savoie, Annecy, France, in 1998. From 1999 to 2023, he has been with Grenoble INP, where he was a Professor of signal and image processing. He is currently a Research Director with INRIA, Grenoble. His research interests include image analysis, hyperspectral remote sensing, data fusion, machine learning and artificial intelligence. He has been a visiting scholar at Stanford University (USA), KTH (Sweden) and NUS (Singapore). Since 2013, he is an Adjunct Professor of the University of Iceland. In 2015-2017, he was a visiting professor at the University of California, Los Angeles (UCLA).  He holds the AXA chair in remote sensing and is an Adjunct professor at the Chinese Academy of Sciences, Aerospace Information research Institute, Beijing.
Dr. Chanussot is the founding President of IEEE Geoscience and Remote Sensing French chapter (2007-2010) which received the 2010 IEEE GRS-S Chapter Excellence Award. He has received multiple outstanding paper awards. He was the Vice-President of the IEEE Geoscience and Remote Sensing Society, in charge of meetings and symposia (2017-2019). He was the General Chair of the first IEEE GRSS Workshop on Hyperspectral Image and Signal Processing, Evolution in Remote sensing (WHISPERS). He was the Chair (2009-2011) and  Cochair of the GRS Data Fusion Technical Committee (2005-2008). He was a member of the Machine Learning for Signal Processing Technical Committee of the IEEE Signal Processing Society (2006-2008) and the Program Chair of the IEEE International Workshop on Machine Learning for Signal Processing (2009). He is an Associate Editor for the IEEE Transactions on Geoscience and Remote Sensing, the IEEE Transactions on Image Processing and the Proceedings of the IEEE. He was the Editor-in-Chief of the IEEE Journal of Selected Topics in Applied Earth Observations and Remote Sensing (2011-2015). In 2014 he served as a Guest Editor for the IEEE Signal Processing Magazine. He is a Fellow of the IEEE, an ELLIS Fellow, a Fellow of the Asia-Pacific Artificial Intelligence Association, a member of the Institut Universitaire de France (2012-2017) and a Highly Cited Researcher (Clarivate Analytics/Thomson Reuters, since 2018).
\end{IEEEbiography}

\end{document}